\pdfoutput=1

\documentclass[11pt]{article}

\usepackage{acl}

\usepackage{times}
\usepackage{latexsym}

\usepackage[T1]{fontenc}

\usepackage[utf8]{inputenc}

\usepackage{microtype}
\usepackage{graphicx}
\usepackage{float}
\usepackage{dblfloatfix} 
\usepackage[skip=2pt]{caption}

\graphicspath{ {plots/} }

\title{Improving Opinion-based Question Answering Systems Through Label Error Detection and Overwrite}

\author{Xiao Yang\thanks{\textbf{  } Correspondence to \texttt{xiaoyangfb@meta.com}}, Ahmed Mohamed, Shashank Jain, Stanislav Peshterliev,\\ {\bf Debojeet Chatterjee},  {\bf Hanwen Zha}, {\bf Nikita Bhalla}, {\bf Gagan Aneja}, {\bf Pranab Mohanty} \\
  Meta Platforms, Inc.\\
  }

\begin{document}
\maketitle

\begin{abstract}
Label error is a ubiquitous problem in annotated data. Large amounts of label error substantially degrades the quality of deep learning models. Existing methods to tackle the label error problem largely focus on the classification task, and either rely on task specific architecture or require non-trivial additional computations, which is undesirable or even unattainable for industry usage. In this paper, we propose LEDO: a model-agnostic and computationally efficient framework for \textbf{L}abel \textbf{E}rror \textbf{D}etection and \textbf{O}verwrite. LEDO is based on Monte Carlo Dropout combined with uncertainty metrics, and can be easily generalized to multiple tasks and data sets. Applying LEDO to an industry opinion-based question answering system demonstrates it is effective at improving accuracy in all the core models. Specifically, LEDO brings $1.1\%$ MRR gain for the retrieval model, $1.5\%$ PR AUC improvement for the machine reading comprehension model, and $0.9\%$ rise in the Average Precision for the ranker, on top of the strong baselines with a large-scale social media dataset. Importantly, LEDO is computationally efficient compared to methods that require loss function change, and cost-effective as the resulting data can be used in the same continuous training pipeline for production. 
\end{abstract}

\section{Introduction}
\label{sec:introduction}

Deep learning model is susceptible to label errors as it can memorize the errors and degenerate the generalization capability \cite{song2022learning}. Here, "label error" refers to the incorrectly assigned target label on data examples for supervised model. 

Label error is a prevalent problem for supervised deep learning model that requires large amount of annotated data to train \cite{song2022learning}. On one hand, high-quality annotated data are expensive and difficult to collect due to the ambiguity of annotation guidelines, annotators not paying enough attention, or even not having sufficient skills to fulfill the task (e.g., low resource language annotation). On the other hand, although Synthetic Data Generation (SDG) methods have been developed to reduce data collection cost, without effective validation, these methods can generate incorrect data that harm model learning \cite{batra2021building}.

Open-domain question answering (QA) aims to address questions based on large-scale unstructured documents.  \cite{zhu2021retrieving}. Collecting high quality annotated data for QA \cite{rajpurkar2018know}, particularly opinion-based QA, is even more challenging. Unlike fact-based QA systems, opinion-based QA is designed to answer questions that do not have a unique factual answer \cite{stoyanov2005multi}. Therefore, answers to opinion-based questions can be subjective and diverse. 

Three key factors contribute to the difficulty of collecting high quality annotated data for opinion-based QA systems for industry. First, judgment for the relevance is subjective and has large variations due to the "opinion-based" nature. Second, for a retriever-reader QA system \cite{chen2017reading} that uses retrieval models such as Dense Passage Retriever (DPR) \cite{karpukhin2020dense}, traditional ways of using BM25 \cite{robertson2009probabilistic} and answer spans to obtain hard negative examples (retrieved documents without the answer span are treated as "hard negatives") can no longer be used, because there is no single unique gold answer to a given question. Lastly, unlike fact-based QA that uses standard knowledge base such as Wikipedia, industry opinion-based QA system is usually based on web contents \cite{he2017dureader}. Texts from web sources are known to be noisy, with low density of information, and informal language \cite{bajaj2016ms}, which inevitably results in elevated annotation error rate and biases.

Many strategies have been proposed to tackle the label error problem. Most of them, however, focus on the classification task \cite{song2022learning}. Meanwhile, model-agnostic, computationally efficient and interpretable solutions are highly desired in industry settings, to reduce development cost and achieve fast scalability.  

\citet{song2022learning} categorized existing methods into 5 major groups: \textit{robust architecture}, \textit{robust regularization}, \textit{robust loss function}, \textit{loss adjustment}, and \textit{sample selection}. While showing theoretical and experimental advancements, the first 3 groups require expensive computation in general \cite{song2022learning, goldberger2017training, zhang2018generalized}, hampering the process of merging into regular training pipelines for industry production usage. Besides, these methods build model robustness implicitly during training, thus providing no explicit signals to guide or improve the annotation process. Loss adjustment methods such as importance reweighting \cite{liu2015classification} or data coefficients \cite{zhang2020distilling} can provide weights for individual examples, but they either rely on manual specification of weighting functions, or require unbiased clean validation sets that are not always available in practice \cite{song2022learning}. These challenges lead us to explore the \textit{sample selection} approaches.

Existing sample selection methods mainly use a multi-network or multi-round iterative approach to guide the selection. Multi-network methods leverage teacher \cite{jiang2018mentornet} or peer \cite{han2018co} networks that are usually constructed based on a similar architecture or underlying data \cite{han2018co, chen2019understanding}. However, this design does not work effectively when there is no clean validation set \cite{han2018co}, or when the error rate is high \cite{chen2019understanding}. Moreover, many sample selection methods \cite{jiang2018mentornet, han2018co, chen2019understanding} rely on the \textit{small-loss} trick, which assumes true-labeled examples have small loss based on the findings that deep learning models usually learn from easy and clean examples first, and then memorize hard or error examples in later learning stages \cite{arpit2017closer}. However, the \textit{small-loss} trick does not work well for the \textit{asymmetric} noise, i.e., when a class is more likely to be mis-labeled as some particular classes rather than equally to all other classes \cite{song2022learning}. These problems are particularly severe for opinion-based QA data collected from web sources due to its noisy and multi-perspective nature, and are further exacerbated in synthetically generated data that use the original data as seeds and has inherited bias from the generative models. 

A promising direction uses a \textit{hybrid} approach that combines sample selection with semi-supervised learning to be able to use the error examples (as unlabeled data) \cite{nguyen2019self}. These methods can bring improved robustness, but are more susceptible to the data and noise types due to the introduced hyper-parameters. Computational cost is also inevitably increased \cite{song2022learning}.

In this paper, we develop a simple yet effective approach -- Label Error Detection and Overwrite (LEDO) that demonstrates to improve all core models: retrieval, reader and ranker in an opinion-based QA system. Different from existing approaches, LEDO uses external models as sentinel model to detect label error whenever possible. This mitigates the issue of inheriting errors when relying on the same underlying noisy set to construct the sentinel model, and also brings in additional information that can benefit the target task. Moreover, instead of requiring the sentinel model's prediction to agree with a given label to indicate it is "correct" \cite{chen2019understanding}, LEDO leverages Monte Carlo Dropout (MCD) \cite{gal2016dropout} that provides more robust signals on the target label space. These design makes LEDO a proper tool to gate synthetic data generation, which otherwise suffers from inherited errors and biases from the seed data and generative models.
Additionally, LEDO is model-agnostic, and computationally efficient compared to methods that require regularization or loss function changes. The resulting data can be used in the same continuous training pipeline and provide guidance to refine the annotation process.



\section{Method}
\label{sec:method}

\subsection{Opinion-based QA System}
An opinion-based question has a multi-perspective nature and cannot be answered by a unique factual answer \cite{stoyanov2005multi}. Here, we study the case of a Community Question Answering (CQA) system that can retrieve and return answers for opinion-based questions that have earlier been asked and answered in a web-based community. Some example questions can be "What is your favorite strawberry smoothie recipe?" or "Recommend me a good seafood restaurant in Seattle."

\subsection{The CQA System Architecture}

\begin{figure*}[ht]
\includegraphics[width=\textwidth]{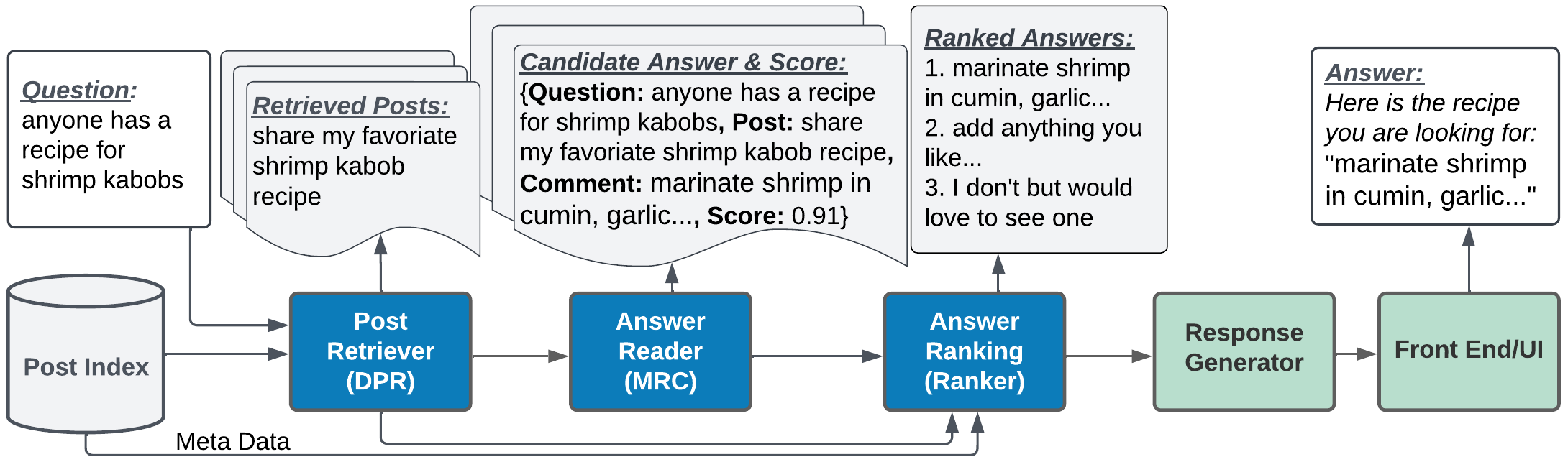}
\caption{The CQA system architecture. Blue blocks are the CQA core models, and green blocks are the front end.}
\label{fig:cqa_architecture}
\vspace{-4mm}
\end{figure*}

The CQA system adopts the modern retriever-reader architecture, and is illustrated in Figure \ref{fig:cqa_architecture}. There are 3 core models. The first is Post Retrieval: we first index an existing post pool that contains $300m$ posts, then use DPR to retrieve the top $k$ most relevant posts for a given question from the user. In the second component, the question, each retrieved post and each comment associated with the post are concatenated together as input into a reader model. The reader uses a Machine Reading Comprehension (MRC) architecture, and is designed to predict whether a given comment associated with a post can answer the user's question, and if it can, identify the answer span within the comment. Since the reader only considers the text features, we use a ranking model that takes the prediction scores from both the retriever and reader, and combines them with metadata such as the number of reactions to the post/comment to rank the final answers to be returned as the last component.

\subsection{Label Error Detection and Overwrite}
\begin{figure*}[h]
\includegraphics[width=16cm]{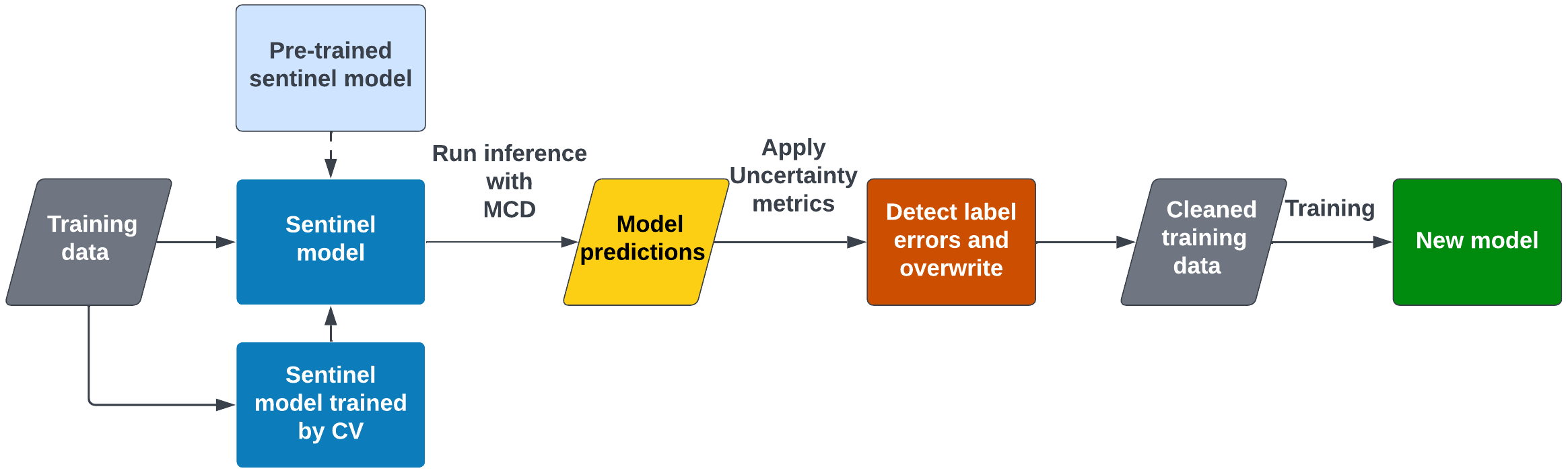}
\caption{The LEDO workflow.}
\label{fig:ledo_workflow}
\vspace{-4mm}
\end{figure*}

The CQA data suffer from label errors heavily due to reasons discussed in Section \ref{sec:introduction}. An analysis on an MRC dev set estimated the error rate to be around $30-50\%$. To address the data quality issue, we develop LEDO, which can automatically detect label error with high precision, and whenever possible, overwrite the label to be the correct one. LEDO workflow is shown in Figure \ref{fig:ledo_workflow}.

LEDO incorporates 3 ingredients: (1) sentinel models that can provide signals to detect label errors; (2) Monte Carlo Dropout that provides a distribution of model predictions; (3) uncertainty metrics to depict the model’s uncertainty on given inputs.

The \textbf{sentinel model} is used to provide signals on whether a label is correct. We prefer to use external models that can make reasonable predictions on the same or similar tasks as the sentinel model whenever possible. Otherwise, we construct it by using cross-validation (CV). \textbf{Monte Carlo Dropout} has been proven to provide approximate Bayesian inference for deep learning models, which can describe uncertainty of existing models without sacrificing computational complexity \cite{gal2016dropout}. It can be implemented conveniently by applying random masks on selected layers of already trained deep learning models. Instead of providing point estimates as normal inference does, it can describe how much uncertainty a subject model has about a given input through its predictive distribution. This information can be summarized in \textbf{uncertainty metrics} such as mean, standard deviation (std), variation ratio etc to help us detect and correct label errors. Concretely, the mean for an MCD distribution is:

\vspace{-4mm}
\begin{equation}
\frac{1}{T} \sum_{t=1}^T \hat{y}(x, W_1^t,... , W_L^t),
\vspace{-2mm}
\end{equation}
where $T$ is the number of forward passes with dropout, $x$ is the model input, $y$ is the output, and $W_i^t$ is the weights in layer $i$ with dropout mask $t$.

\subsection{Generate Negative Synthetic DRP Data}
\label{sec:synthetically_generated_data}

Effective DPR training relies on the inclusion and quality of hard negative examples \cite{karpukhin2020dense}. However, traditional methods of automatically obtaining hard negatives based on the answer span cannot be used for opinion-based QA as discussed in Section \ref{sec:introduction}. Hence, we leveraged an SDG strategy to create nuanced negative examples, and employed LEDO to remove noise that can hurt model training from this large-scale synthetic set. 

\citet{batra2021building} introduced an approach of creating nuanced synthetic negative data by performing mask-filling using a domain adapted fine-tuned BART \cite{lewis2019bart} model. We took each question as a seed and designed "rules" to replace salient parts from the question so that the resulting sentence has a different semantic meaning, thus making it a nuanced negative instance. 
The BART model is first fine-tuned on in-domain CQA data with the objective of reconstructing the masked input sequence. Spacy Parser \cite{spacy2} provides us access to nouns, compound nouns, verbs etc and their dependencies. We designed 4 set of rules using the Spacy parser to find candidate phrases to mask. For example, for the seed question "Can someone recommend me a good \textit{recipe} for alfredo sauce", one of the rules identified to mask the word "recipe", and the fine-tuned mask-filling BART model kept the original sentence structure while replacing the word "recipe" with a different meaning in-domain word "substitute", therefore created a negative example "Can someone recommend me a good \textit{substitute} for alfredo sauce". Details on how the rules are designed can be found in Appendix \ref{sec:synthetic_data_generation_rules}. 


\section{Experiments and Results}
\label{sec:results}
In this section, we elaborate experiments for applying LEDO to the DPR, MRC and ranker in CQA. 

\subsection{Data}
Training data come from 2 sources: human-annotated and synthetically generated. For evaluation, we collected additional annotation on a subset of the annotated data to use as clean test sets.

Collecting \textbf{human-annotated data} for the CQA system involves 4 annotation tasks. The first task is \textit{question generation}. It samples a set of posts from the pool, and asks the annotators to write questions that can be answered using those posts. This is followed by \textit{post retrieval} in which we use a baseline retriever (e.g., BM25) to retrieve a number of posts for each human-generated question, and ask the annotators to label whether a post is relevant, neutral or irrelevant to that question. The third task takes the posts that are labeled as "relevant" and asks the annotator to provide a binary label on whether this post and its associated comment can answer the corresponding human-generated question. If it can, then mark the span within the post text that could be the answer. Last, a subset of the identified answers are sent to a different set of annotators to mark a score between 1 to 5 to indicate how good the answer is (with 5 being the best and 1 being the worst). This subset of data is subsequently used to train the ranker. The question-post pairs with their ternary labels from the first and second tasks are used to train the DPR model. The binary labeled data along with answer spans from the third task is used to train the MRC model. All of these human-annotated data become a major source of training data for the DPR, MRC and ranking tasks.

To cope with lack of hard negative DPR data, we created \textbf{synthetic data} as discussed in Section \ref{sec:synthetically_generated_data}.

\subsection{Dense Passage Retrieval}
The DPR model was originally trained using $96k$ human-annotated data. We created a $1.2m$ synthetic set as nuanced negative examples\footnote{In each training iteration, a negative example is randomly sampled from the available hard negative examples.}. 


We use a RoBERTa entailment model fine-tuned on the MNLI corpus \cite{liu2019roberta} as the sentinel model to detect and filter out falsely labeled examples from both the annotated and synthetic data. 
Here, we first run inference using the entailment model with MCD on the subject data. For positive data, we remove examples if the entailment model predicts "contradiction" with a mean score above $t_1$ and std below $s_1$; likewise, for negative data, we reject examples with mean "entailment" score above $t_2$ and std below $s_2$. The threshold values are tuned on a dev set and included in Appendix \ref{sec:hyper_parameters_for_LEDO}. \footnote{We didn't overwrite the DPR labels since the clean dev set is quite small so we only use it to tune the thresholds.} About $7\%$ of the annotated data and $59\%$ of the synthetic data were removed in this process. 

We tried 2 ways of injecting the synthetic data: \textbf{blend}: directly augment the existing training data; \textbf{pre-train}: use "blend" to pre-train the model, and then fine-tune with LEDO cleaned annotated data. Results from the DPR experiments are shown in Table \ref{tab:dpr_results}. 
Applied separately SDG and LEDO brings around $0.6$ Mean Reciprocal Rank (MRR) gain, with LEDO showing another $0.7$ drop in Avg. Rank, indicating LEDO improves more on top candidates. Combining SDG and LEDO with pre-training captures the largest improvement: $1.04\%$ increase on MRR and $1.09$ drop in Avg. Rank. This reflects the synergy between SDG and LEDO: SDG creates large-scale negative data required by DPR which is costly to obtain with human annotation; LEDO filters out noise from this large-scale set and produces the best result by providing a cleaner version of both the annotated and synthetic data. 

\begin{table}[h]
\centering
\begin{tabular}{lcc}
\hline
\textbf{DPR Experiment} & \textbf{MRR} & \textbf{Avg. Rank} \\
\hline
     Baseline & 84.06 & 4.43 \\ 
     + SDG & 84.73 & 4.26 \\
     + LEDO & 84.69 & 3.7 \\
     + SDG \& LEDO, blend & 84.83 & 3.76 \\
     + SDG \& LEDO, pre-train & \textbf{85.2} & \textbf{3.34} \\
\hline
\end{tabular}
\caption{LEDO improves DPR model quality by filtering out noise from both the annotated and synthetic data. 
}
\label{tab:dpr_results}
\vspace{-4mm}
\end{table}

\subsection{Machine Reading Comprehension}
The MRC model is based on a standard RoBERTa \cite{liu2019roberta} encoder and uses a binary classifier to decide whether a comment, from a retrieved post, contains an answer for the given question. If the comment does answer the question, the model additionally extracts the corresponding answer span from the comment. \footnote{Here the answer span head is used as an auxiliary task to help the MRC model learn and provide features for the ranker. We didn't use it to generate the final user-facing results.}

Since existing MRC models are mainly trained with fact-based data such as Wikipedia \cite{liu2019neural}, these models do not work well on CQA questions that come from web contents. Hence, we use cross-validation to build the sentinel models, and again use mean and std as the uncertainty metrics. Specifically, when the mean and std of the MCD predictions are both below the pre-specified thresholds $t_1, s_1$ and the example was marked as "positive", we consider it to be a false positive and overwrite the label to be "negative". Similar steps can apply to false negative predictions with the mean criteria changing to be above a threshold $t_2$. The thresholds are selected based on a clean dev set, and their values are included in Appendix \ref{sec:hyper_parameters_for_LEDO}.
About $7\%$ out of the total $750k$ labeled examples were corrected from this process with $> 80\%$ precision. We then use the cleaned data to retrain the MRC model. The results are shown in Table \ref{tab:mrc_results}.

\begin{table}[h]
\centering
\begin{tabular}{lcc}
\hline
 \textbf{MRC Experiment} & \textbf{F1} & \textbf{PR AUC} \\
\hline
 Baseline & 78.6 & 76 \\
 LEDO     & 79.6 & 77.5 \\
\hline
\end{tabular}
\caption{Correcting errors in the MRC data leads to $1\%$ and $1.5\%$ F1 and PR AUC improvement respectively.}
\label{tab:mrc_results}
\vspace{-4mm}
\end{table}

\subsection{Ranker}
The ranker uses a 2-layer neural network with RELU activation. We optimize the ranker using the Lambda Rank objective which is based on pairwise classification of a 5-point ranking set. The ranker takes inputs from the DPR and MRC models along with additional meta signals and produces a final order for posts and comments.

\begin{table}[h]
\centering
\begin{tabular}{lc}
\hline
\textbf{Ranker Experiment} & \textbf{Avg. Precision} \\
\hline
  Baseline & 77.9 \\
  LEDO     & 78.8 \\
\hline
\end{tabular}
\caption{The Avg. Precision of the ranker is improved by $0.9\%$ after training with the LEDO selected set.}
\label{tab:ranker_results}
\vspace{-2mm}
\end{table}

The ranker model is not naturally a good choice for sentinel model as scores across examples are not comparable. Also, since we care more about \textit{major} errors such as a rank 1 answer marked as rank 5 than a rank 4 marked as 5, we trained a RoBERTa \cite{liu2019roberta} classifier with CV as the sentinel model using the 5-point ranking set to classify each given answer as bad (score 1 and 2), neutral (score 3) or good (score 4 and 5). Then, we used the sentinel classifier to run inference with MCD over the ranking data, and rejected original "good" examples when the $q_1$ quantile is "neutral" or "bad". Likewise, original "bad" or "neutral" examples with a $q_2$ quantile being "good" were also rejected. We use quantile as the uncertainty metric here due to its interpretability and ease of use as there is no clean dev set to tune the thresholds. This process removes about $10\%$ data from the $10k$ ranking set. Results for training the new ranker are shown in Table \ref{tab:ranker_results}.

\subsection{End-2-End Evaluation}
Besides component-wise evaluation, we also conducted an end-2-end (E2E) evaluation to measure the user-facing accuracy. The raters view the question, the retrieved posts/comments and the returned answers, and assess the answers on a 5-point scale, with 1 being the worst and 5 being the best. Questions with an answer scored 4 or 5 are considered "helpful". In order to guarantee high precision, we set a confidence threshold $\tau$ to control whether a post/comment will be returned to the user as answer. We define \textbf{recall} to measure how many questions are answered, i.e., having confidence scores passing $\tau$, and \textbf{precision} as the number of questions having a score $\geq 4$ among all questions that the system returns an answer (recalled). Results from the E2E evaluation are presented in Table \ref{tab:e2e_results}. 

\begin{table}[h]
\centering
\begin{tabular}{lcc}
\hline
\textbf{E2E Experiment} & \textbf{Rel. Recall Change} \\
\hline
LEDO on DPR               & +8.55 \\
LEDO on MRC $\&$ Ranker   & +5.57 \\
\hline
\end{tabular}
\caption{E2E evaluation measures the user-facing accuracy. By locking the precision at $80\%$, the improvement from the DPR, MRC and ranker experiments lead to $8.55\%$ and $5.57\%$ relative gain on the recall. The \textit{Relative Recall Change} is calculated as 
$\mbox{absolute recall change} / (1 - \mbox{baseline recall}).$}
\label{tab:e2e_results}
\vspace{-4mm}
\end{table}


\subsection{Result Analysis}
\begin{figure}[h]
\includegraphics[width=7.7cm]{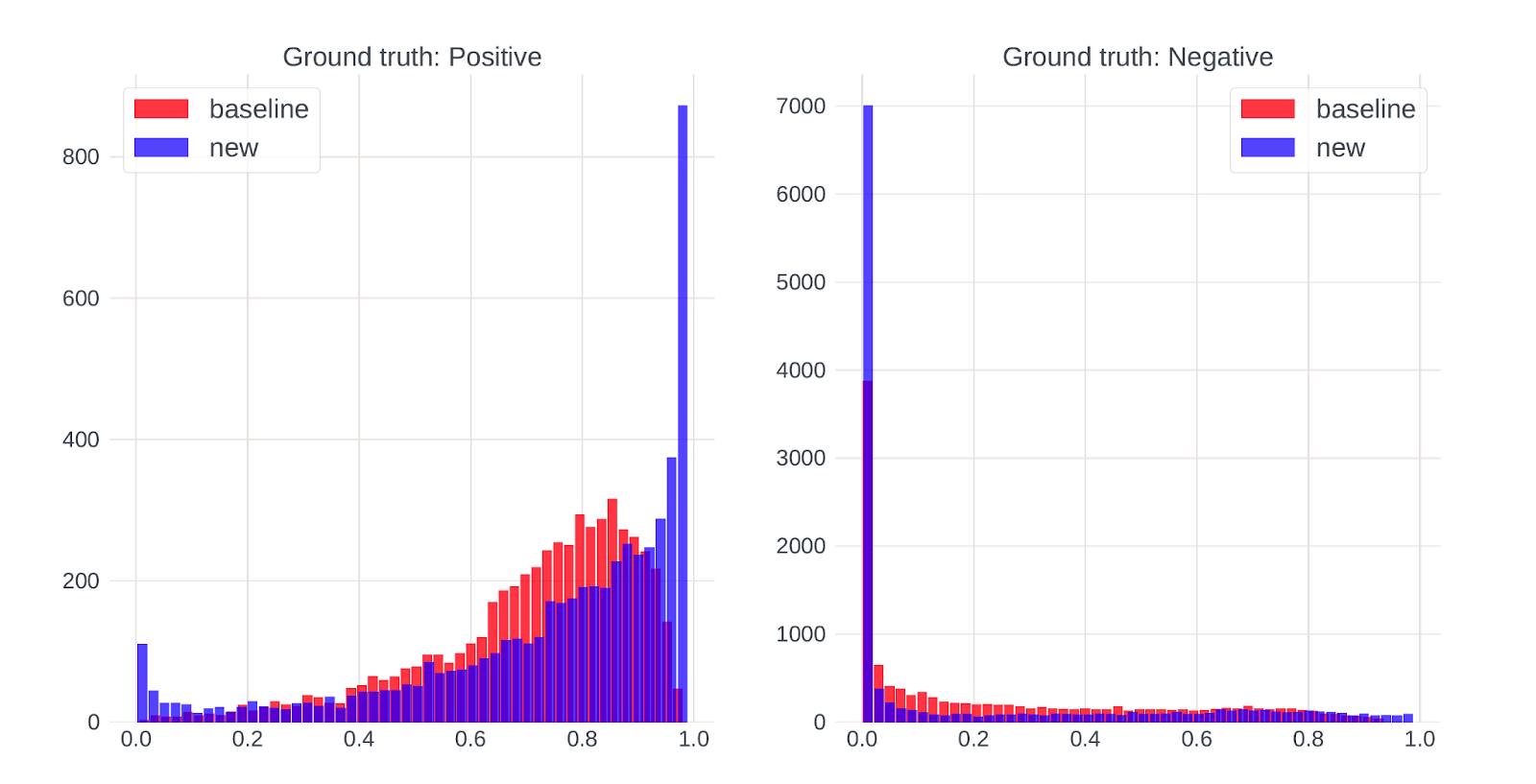}
\caption{The new MRC model trained with clean data becomes more confident in the correct predictions.}
\label{fig:prediction_probability}
\vspace{-4mm}
\end{figure}

Compared to the baseline MRC model, the new model is observed to push the probabilities on the positive data to be closer to 1 and the negative data to be closer to 0 as shown in Figure \ref{fig:prediction_probability}. This indicates the new model becomes more confident on its correct predictions, and the decision boundary is refined after label errors are removed.

\begin{figure}[h]
\includegraphics[width=7.7cm]{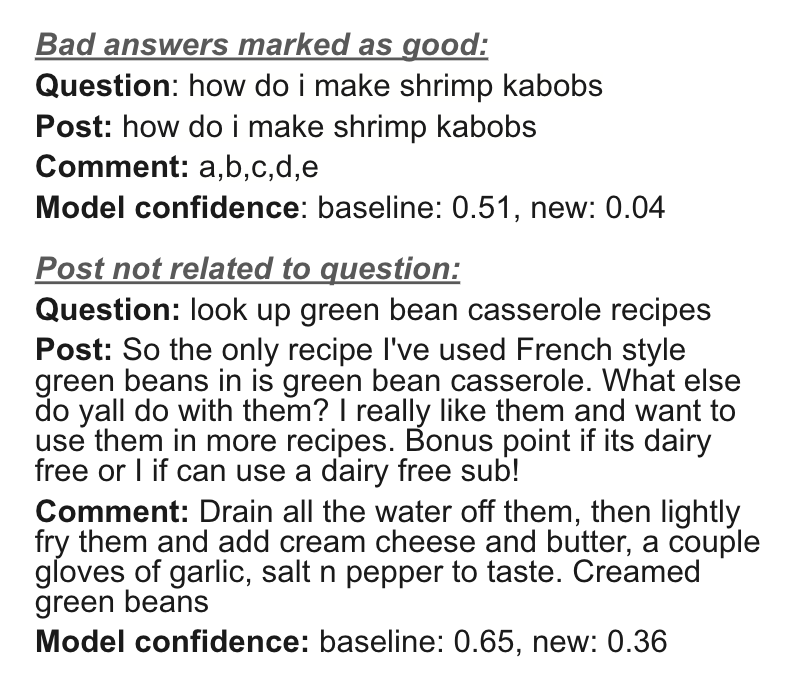}
\caption{The baseline MRC model tends to predict positive on any answer when the post and question are very similar. The new model corrects this behavior and gives a low score to an artificially made answer "a,b,c,d,e" as shown in the first example. The second example illustrates a case when the baseline model overweighs the phrase "green bean casserole" and ignores the "else" condition. The new model resolves this error by giving a low score to the seemingly correct comment.}
\label{fig:improved_examples_mrc}
\vspace{-2mm}
\end{figure}

\begin{figure}
\includegraphics[width=7.5cm]{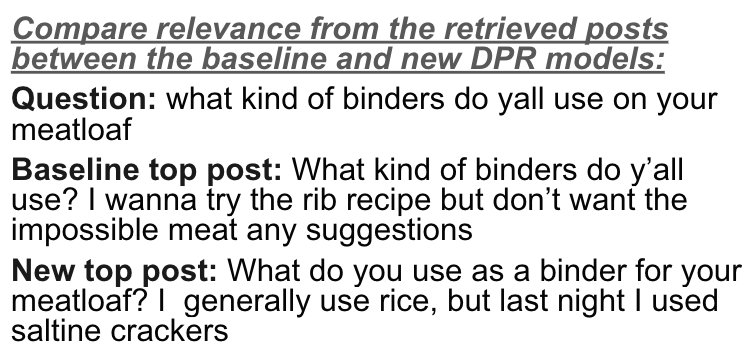}
\caption{The new DPR model has improved relevance for the retrieved posts compared to the baseline model.}
\label{fig:improved_examples_dpr}
\vspace{-4mm}
\end{figure}

We observe the improved examples with the new MRC model mainly come from correcting previous mistakes sitting near the decision boundary, as illustrated by the examples in Figure \ref{fig:improved_examples_mrc}. For the DPR model, improvement is from enhanced relevance of the retrieved posts as shown in Figure \ref{fig:improved_examples_dpr}.

The computational cost involved in applying LEDO comes from running MCD inference with $T$ forward passes. In our experiments $T=10$ shows good performance. This can be done offline hence does not incur any runtime latency cost. Besides, it is less expensive and widely-applicable compared to loss function change during the training process.

\section{Conclusion and Future Work}
\label{sec:conclusion}
High quality annotated data are expensive to obtain at scale. It is even more challenging for industry opinion-based QA system due to the subjectivity of the answers and noisy nature of the contents. We developed LEDO: a versatile and computationally efficient framework that can automatically detect and correct label errors for supervised data, and demonstrated it effectively improves the retriever, reader and ranker in a large-scale industry opinion-based QA system. The design of LEDO provides a control knob on how much uncertainty we want to include in the identified examples: when confident, we can directly overwrite; otherwise, we request additional annotators' view to get the correct label. It would be interesting to see the value added from this re-annotation. Moreover, extending LEDO to tasks such as generative models is also an interesting topic to investigate.


\bibliography{anthology,custom}
\bibliographystyle{acl_natbib}

\appendix
\section{Appendix}
\subsection{Spacy-based Rules for Generating Nuanced Negative Data}
\label{sec:synthetic_data_generation_rules}

\begin{figure*}[ht]
\includegraphics[width=16cm]{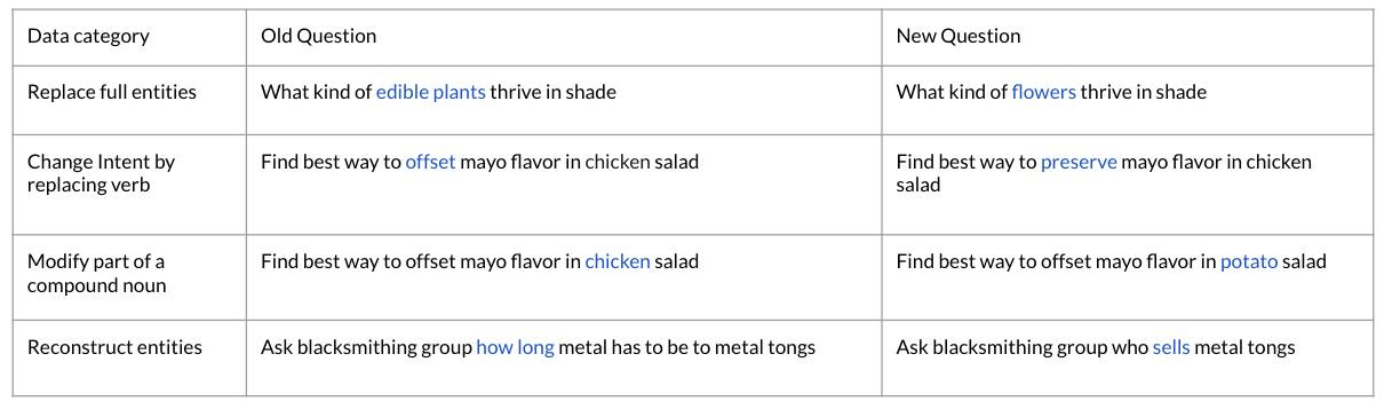}
\caption{Synthetically generated nuanced negative examples using Spacy-based rules}
\label{fig:sdg_4rules}
\end{figure*}

We designed 4 rules using Spacy to create masked questions for generating nuanced negative examples for the DPR model:
\begin{enumerate}
    \item Modify part of compound noun by identifying and replacing the first part of compound noun.
    \item Replace full entities by identifying and replacing nouns and proper nouns.
    \item Change intent by identifying and replacing verbs.
    \item Reconstruct sentence by preserving the entities and masking out everything else to create sentences with semantically different meaning but same entities.
\end{enumerate}
Figure \ref{fig:sdg_4rules} shows some example negative questions created based on the 4 rules.

\begin{table*}[ht]
\setlength\tabcolsep{2em}
\centering
\begin{tabular}{lccc}
\hline
\textbf{Model} & \textbf{Dropout Rate} & \textbf{T} & \textbf{Uncertainty Metrics} \\
\hline
DPR    & $0.1$ & $10$ & $t_1=0.75, s_1=0.2, t_2=0.7, s_2=0.2$ \\
MRC    & $0.1$ & $10$ & $t_1=0.3, s_1=0.15, t_2=0.75, s_2=0.15$ \\
Ranker & $0.1$ & $10$ & $q_1=90\%, q_2=10\%$ \\
\hline
\end{tabular}
\caption{Hyper-parameters used in the LEDO experiments. For the DPR and MRC experiment, mean and std were used as uncertainty metrics. The thresholds are tuned based on clean dev test sets. For the ranker experiment, we used quantile as the uncertainty metric, and picked the thresholds based on a small number of manually checked examples.}
\label{tab:ledo_hyperparameters}
\end{table*}

\subsection{Hyper-parameters Used in the LEDO Experiments}
\label{sec:hyper_parameters_for_LEDO}

Table \ref{tab:ledo_hyperparameters} includes the hyper-parameters used in the LEDO experiments for the DPR, MRC and ranker models. For all 3 experiments, the dropout rate was set to equal to the rate used during training. And the number of stochastic predictions are set to 10 since we found further increase does not bring additional improvement in our experiments.

\end{document}